\documentclass{article}

\PassOptionsToPackage{numbers}{natbib}

\usepackage[final]{tackling_climate_workshop_style}

\usepackage[utf8]{inputenc} %
\usepackage[T1]{fontenc}    %
\usepackage{hyperref}       %
\usepackage{url}            %
\usepackage{booktabs}       %
\usepackage{amsfonts}       %
\usepackage{nicefrac}       %
\usepackage{microtype}      %
\usepackage{amsmath}
\usepackage{commath}
\usepackage{amssymb}
\usepackage{graphicx}
\usepackage{caption}
\usepackage{subcaption}
\usepackage{booktabs}
\usepackage{wrapfig}
\usepackage{dsfont}
\title{
Optimizing ship detection efficiency in SAR images
}

\author{%
  Arthur Van Meerbeeck \\
    \texttt{arthurvanmeerbeeck@gmail.com} \\
  \And
  Jordy Van Landeghem \\
  \texttt{jordy.vanlandeghem@kuleuven.be} \\
   \And
   Ruben Cartuyvels\thanks{Corresponding author.}
   \\
  \texttt{ruben.cartuyvels@kuleuven.be} \\
   \And
  Marie-Francine Moens \\
  \texttt{sien.moens@kuleuven.be}
  \\
 \And
  {\normalfont Department of Computer Science} \\
  KU Leuven, 3001 Leuven (Belgium) \\
}

\begin{document}

\maketitle

\begin{abstract}

    The detection and prevention of illegal fishing is critical to maintaining a healthy and functional ecosystem. 
    Recent research on ship detection in satellite imagery has focused exclusively on performance improvements, disregarding detection efficiency. 
    However, the speed and compute cost of vessel detection are essential for a timely intervention to prevent illegal fishing. 
    Therefore, we investigated optimization methods that lower detection time and cost with minimal performance loss. 
    We trained an object detection model based on a convolutional neural network (CNN) using a dataset of satellite images. 
    Then, we designed two efficiency optimizations that can be applied to the base CNN or any other base model.
    The optimizations consist of a fast, cheap classification model and a statistical algorithm. 
    The integration of the optimizations with the object detection model leads to a trade-off between speed and performance. 
    We studied the trade-off using metrics that give different weight to execution time and performance. 
    We show that by using a classification model the average precision of the detection model can be approximated to 99.5\% in \textpm44\% of the time or to 92.7\% in \textpm25\% of the time.
\end{abstract}

\section{Introduction}

Biodiversity conservation and climate change are significant and current 
global issues. 
A major contemporary problem in biodiversity conservation is overfishing, which often results from Illegal, Unreported and Unregulated (IUU) fishing. 
It is estimated that IUU fishing is responsible for 30\% of all fishing in the world, with an estimated economic loss of up to \$23 billion per year \cite{IUU}. 
Not only do illegal activities threaten marine ecosystems, but the resistance of many marine species to climate change is compromised by overfishing, which increases the vulnerability of marine fisheries production to ocean warming. 
Furthermore, ongoing warming will impede efforts to recover overfished populations \cite{Climatechange}. 
Satellite imagery can be used to detect vessels on sea and to determine if these vessels are conducting illegal activities. 

Synthetic Aperture Radar (SAR) images are created using satellites that emit radio waves towards the earth and capture the waves reflected back by objects \cite{SAR_tutorial}.
When a ship is detected on these images, it can be determined whether a corresponding Automatic Identification Signal (AIS), which all ships should transmit every few seconds, can be linked to that ship.
If this is not the case, the ship could possibly be conducting illegal activities \cite{SARAIS}. 
Some satellites can image about 7 million square miles per day and thus various machine learning algorithms are used nowadays to detect ships in these SAR images \cite{Skylight}. 
However, current models are not yet able to process the SAR data both efficiently and with high accuracy, which results in high resource costs and which could impede their use in real-time systems. 
Therefore, there is a need for models that can quickly and accurately detect ships in SAR images o allow the Coast Guard to take timely and efficient action when the models detect illegal fishing \cite{Speed}. 
The speed and performance of machine learning models are often negatively correlated and, therefore, it is necessary to make an appropriate trade-off for this.

\begin{wrapfigure}{r}{0.7\textwidth}
    \centering
    \includegraphics[width=1.0\linewidth]{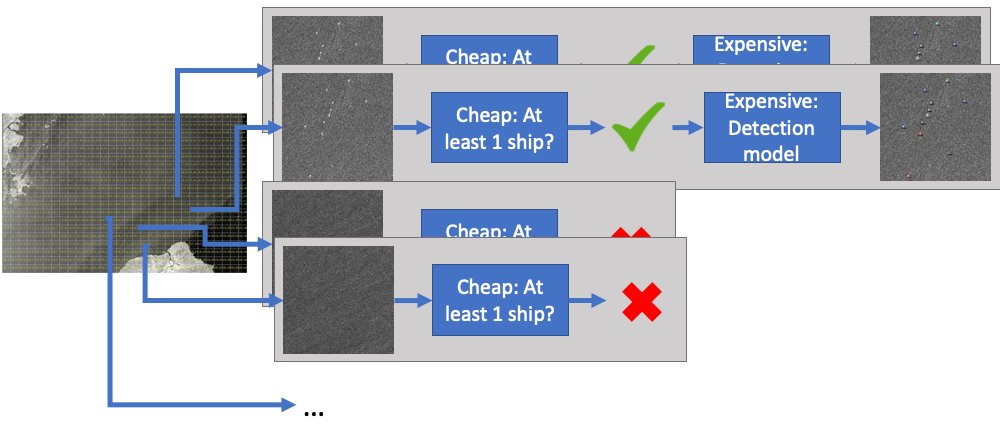}
    \caption{Overview of the optimizations and the detection model. If in a first step, a fast algorithm predicts that no ship is present, the expensive detection model is skipped.}
    \label{fig:overview}
\end{wrapfigure}

We investigated optimizations for machine learning models, agnostic as to how the base model functions, that detect ships in SAR images to make the best possible trade-off. 
These optimizations consist of a fast classification model and a statistical algorithm. 
Their goal is to efficiently determine as a first step whether at least one ship is present in an SAR image before applying the expensive base detection model. Figure \ref{fig:overview} gives an overview.

\section{Related Work}
The launch of the first SAR satellite in the United States in 1978 led to the emergence of several SAR ship detection methods \cite{Satellite}. These initial methods are based on traditional detection methods that manually design features \cite{Trad1,Trad2,Trad3}. 
These traditional methods proved too slow, required complex manual work and were inaccurate \cite{SSDD}. 
The rise of neural networks led to the use of advanced object detection techniques due to their higher accuracy and lesser need for human intervention. 
This was also the case for the SAR ship detection community: the proliferation of SAR data \cite{xview} in recent years has contributed to the increase in the use of deep learning (DL) in SAR ship detection \cite{CNN1,CNN2,CNN3}. In these models, the traditional features (such as gray level, HOG, etc.) are replaced by features produced by CNNs.
The SAR dataset used in this study is the Large-Scale SAR Ship Detection Dataset-v1.0 (LSSSDD) \cite{LSSSDD}. The dataset contains 15 SAR images from the Sentinel-1 satellite \footnote{\url{https://scihub.copernicus.eu}}, with a size of $16000 \times 24000$ pixels, divided into $900$ sub-images of size $800 \times 800$.

\section{Efficiency optimizations}
A large percentage of the SAR sub-images do not contain ships and so time is lost by performing object detection on them. The goal of our optimizations is to perform detection on as few images as possible with minimal loss of performance. 
To achieve this, we use two models that efficiently determine for each sub-image whether a ship is present or not, before applying the more expensive ship detection model that locates each ship. 
If the models conclude that no ship is present on an image, the image is not fed to the detection model. 
A trade-off has to be made between inference time and performance, by choosing the positive classification threshold of the efficient first-stage classification.

\paragraph{Fast classification with small CNN}

This optimization implements a binary classification model with a CNN. 
The CNN produces a score $s$ between -1 and 1, with a score lower than the chosen threshold $t^\text{clf}$ indicating that at least one ship is present, a higher score that no ship is present.
Since the goal of optimization is saving time, the small and efficient MobileNetV2 \cite{MobileNetV2} architecture was used. 
$I$ is a SAR sub-image of size $800 \times 800$.
\begin{equation}
    f_\text{clf}(I,{t}^\text{clf}) = \operatorname{CNN}(I) = s^\text{clf} \; \begin{cases}
    I \text{ contains a ship} & ~\text{if}~ s^\text{clf}\leq{t}^\text{clf} \\ 
    I \text{ does not contain a ship} & ~\text{if}~ s^\text{clf} >{t}^\text{clf}
    \end{cases}
\end{equation}

The MobileNetV2 model minimizes inference time with minimal loss of performance by using a new type of layer module. 
Initially, weights trained on the ImageNet dataset \cite{Imagenet} are loaded in. Then, the final layers of the model are retrained on the LSSSDD until the validation loss converges.

\paragraph{Correlation in ship presence between neighboring images}
Another way to avoid running object detection is to predict ship presence in a given sub-image based on the presence of ships in neighboring sub-images. In this optimization, object detection is first conducted on a subset of the sub-images. The images on which object detection is initially performed are chosen via two patterns shown in Figure \ref{fig:patronen}.
The optimization with the checkers and $\alpha$ pattern are respectively noted as $f_\text{cor-checkers}$ and $f_\text{cor-$\alpha$}$.
The patterns check respectively 50\% and 25\% of all the sub-images in the initial step. 
Afterward, using that information, the correlation algorithm predicts for each of the remaining sub-images whether a ship is present.
This is done by taking the weighted average of a ship presence indicator of neighboring images on which ship detection was conducted. 
If $s^{cor}$ is greater than a chosen threshold ${t}^\text{cor}$, the image is fed to the object detection model. 
The time to calculate $s_{cor}$ is negligible in comparison to the time necessary to perform object detection.
\begin{equation} \label{eq:correlation}
    f_\text{cor}(I,K,w,{t}^\text{cor}) = 
    \sum_{j \leq K}
    \sum_{i\in N_j} w_{j} \mathds{1}_i = s^\text{cor} \; \begin{cases}
    I \text{ contains a ship} & ~\text{if}~ s^\text{cor}\geq{t}^\text{cor} \\ 
    I \text{ does not contain a ship} & ~\text{if}~ s^\text{cor}<{t}^\text{cor}
    \end{cases}
\end{equation}
In this equation, $N_j$ is the set of neighbours that are $j$ tiles away, $w_j$ is the weight given to all neighboring tiles that are $j$ tiles away, and $\mathds{1}_i$ indicates whether neighbor $i$ contains at least one ship or not.
Neighbours that are 1 tile away get the largest weight $w_1$, neighbours that are 2 tiles away a smaller weight, and so on. 

\section{Experiments}

An overview of the split of the data set is given in Table \ref{tab:datasets}. 
We used a Faster R-CNN model with a ResNet-50 backbone to detect ships in the SAR images \cite{ren2015frcnn,he2016resnet}. 
We start training from pre-trained weights for object detection on MSCOCO \cite{lin2014coco}, and retrain on the LSSSDD dataset with use of the Detectron2 library\footnote{\url{https://github.com/facebookresearch/detectron2}} until the validation loss converges.

\subsection{Results}
\begin{wraptable}{l}{0.62\textwidth}
\addtolength{\tabcolsep}{-3pt}
\footnotesize
\centering
\begin{tabular}{ c  c c  }     
    \toprule

    {Model} & {Total Time (s)} & {AP} \\
    \midrule
    Baseline & 810.84 & 0.711  \\ 
    $f_\text{clf}(I,t^\text{clf}_1)$ & 283.93 & 0.693 \\
    $f_\text{clf}(I,t^\text{clf}_2)$ & 362.70 & 0.706 \\
    $f_\text{cor-$\alpha$}(I,K_1,w_1,t^\text{cor}_1)$ & 405.41 & 0.616 \\
    $f_\text{cor-$\alpha$}(I,K_2,w_2,t^\text{cor}_2) + f_\text{clf}(I,t^\text{clf}_3)$ & 202.71	&	 0.638 	 \\
    
    \bottomrule
\end{tabular}

\caption{Total time and average precision for five SAR images.
}
\label{tab:Results}
\end{wraptable}

Table \ref{tab:Results} shows that the baseline Faster R-CNN model achieves an Average Precision (AP) of 0.711 in a total of 811s (including load and detection time). 
All optimizations drastically reduce the total time, while losing limited AP.
When applying the optimizations, time savings can be weighed against performance retention.
For this purpose, the $\digamma_{\beta}$ score is calculated between the AP of the model and the time savings compared to the baseline model. The time gain is equal to 1 - the relative time (RT) compared to the baseline model ($ \text{RT} = \frac{\text{Total time with optimization}}{\text{Total time of baseline}}$). 
The smaller $\beta$, the more weight the AP gets in the $\digamma_{\beta}$ score. 

\begin{equation}
    \digamma_{\beta} = (1+{\beta}^{2}) \cdot \dfrac{AP \cdot (1-RT)}{{\beta}^{2} \cdot AP + (1-RT)}
\end{equation}

\begin{figure}
    \centering
    \includegraphics[width=0.8\linewidth]{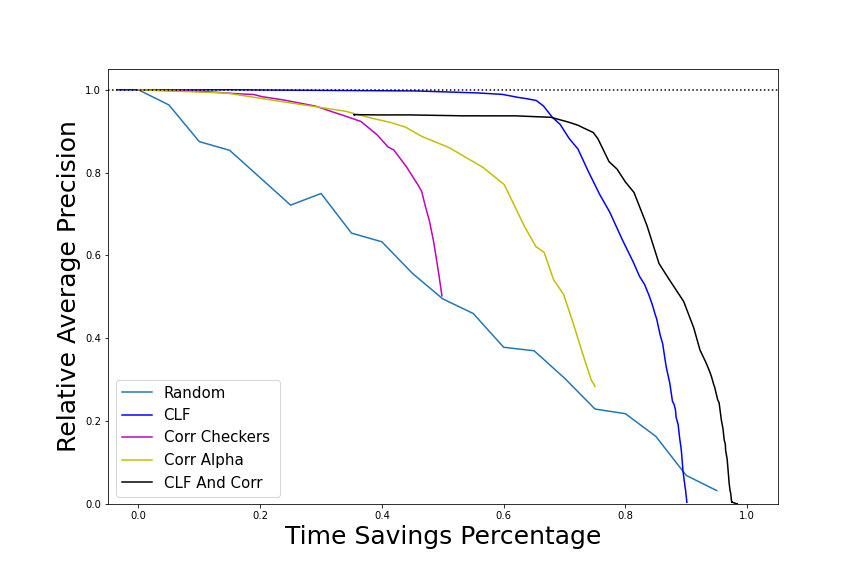}
    \caption{Relative AP and percentage of time saving for the optimizations and an algorithm that deletes random sub-images from the testset. 
    }
    \label{fig:GraphAll}
\end{figure}

\paragraph{Classification model}
A fast CNN first predicts which images contain a ship.
The classification time of all sub-images of one SAR image is approximately 6s. 
The best results are given in Table \ref{tab:Results}. 
With $t^\text{clf}_1 = 0$ the model achieves an AP of 0.693 in 35\% of the time of the baseline model, resulting in a $\digamma_{1}$ score of 0.673. 
This optimization produces the highest overall $\digamma_{0.25}$ score (AP has higher weight) with $t^\text{clf}_2 = 0.2$, with $\digamma_{0.25} = 0.695$. 
This model achieves an AP of 0.706 in 44\% of the time.

\paragraph{Correlation algorithm}
This optimization determines whether to run detection on a sub-image based on the presence of ships in neighboring images, following eq. \eqref{eq:correlation}. 
The calculation of the correlation score is negligible compared to the object detection time of the sub-images. 
The highest $\digamma_{1}$ score achieved is equal to $0.553$ with $K = 2$, $w_1 = 1, w_2 = 0.1$ and $t_{1}^\text{cor} = 0.4375$ as seen in \ref{tab:Results}. 
The highest $\digamma_{0.25}$ score is equal to $0.637$. This optimization, therefore, does not surpass the $\digamma_{\beta}$ of the classification optimization.

\paragraph{Combination}
Both optimizations are combined by first dividing the SAR sub-images according to the chosen correlation pattern and, before performing detection, first predicting using the fast CNN whether the sub-image contains a ship.
As a result, the detection model only performs detection on the sub-images that are classified as containing a ship by both optimizations. 
This combination produces a $\digamma_{1}$ score (time savings and AP have equal weight) of 0.688, higher than the individual classification optimization.
It achieves an AP of 0.638 in only 25\% of the time of the baseline model. 
This is shown in Table  \ref{tab:Results} were $K = 2$, $w_1 = 1, w_2 = 0.5$, $t^\text{cor}_2 = 0.25$ and $t^\text{clf}_3 = 0$.

Figure \ref{fig:GraphAll} shows the time savings vs. the AP as a curve per optimization, with the points on the curve corresponding to different thresholds $t$.
As expected from the $\digamma_{\beta}$ scores, the classification optimization retains performance best (curve with highest y-coordinates).
Both optimizations perform better than an algorithm that randomly removes images from the test set. 
It is also visible that the combination of both optimizations outperforms individual optimizations in terms of time savings (curve with highest x-coordinates).

\paragraph{Conclusion}
In this paper, we studied vessel detection in SAR images to counter illegal fishing. 
We introduced and tested two optimization techniques as a first-stage filter to make the detection as efficient as possible. 
From the experiments, we concluded that a classification model based on MobileNetV2 yields the best results when more weight is given to performance retention: approximating the baseline AP to 99.5\% in 44\% of the time.
If the reader gives equal weight to time savings and performance, the combination of the correlation- and classification-optimization is best, approximating the baseline AP to 92.7\%, in only 25\% of the time.
We hope this study makes clear the importance of detection efficiency, and paves the way for more efficiency improvements.

\clearpage

\bibliography{tackling_climate_workshop}

\clearpage
\appendix

\section{Appendix}

\subsection{LSSSDD}

\begin{figure}[h]
    \centering
    \includegraphics[width=10cm]{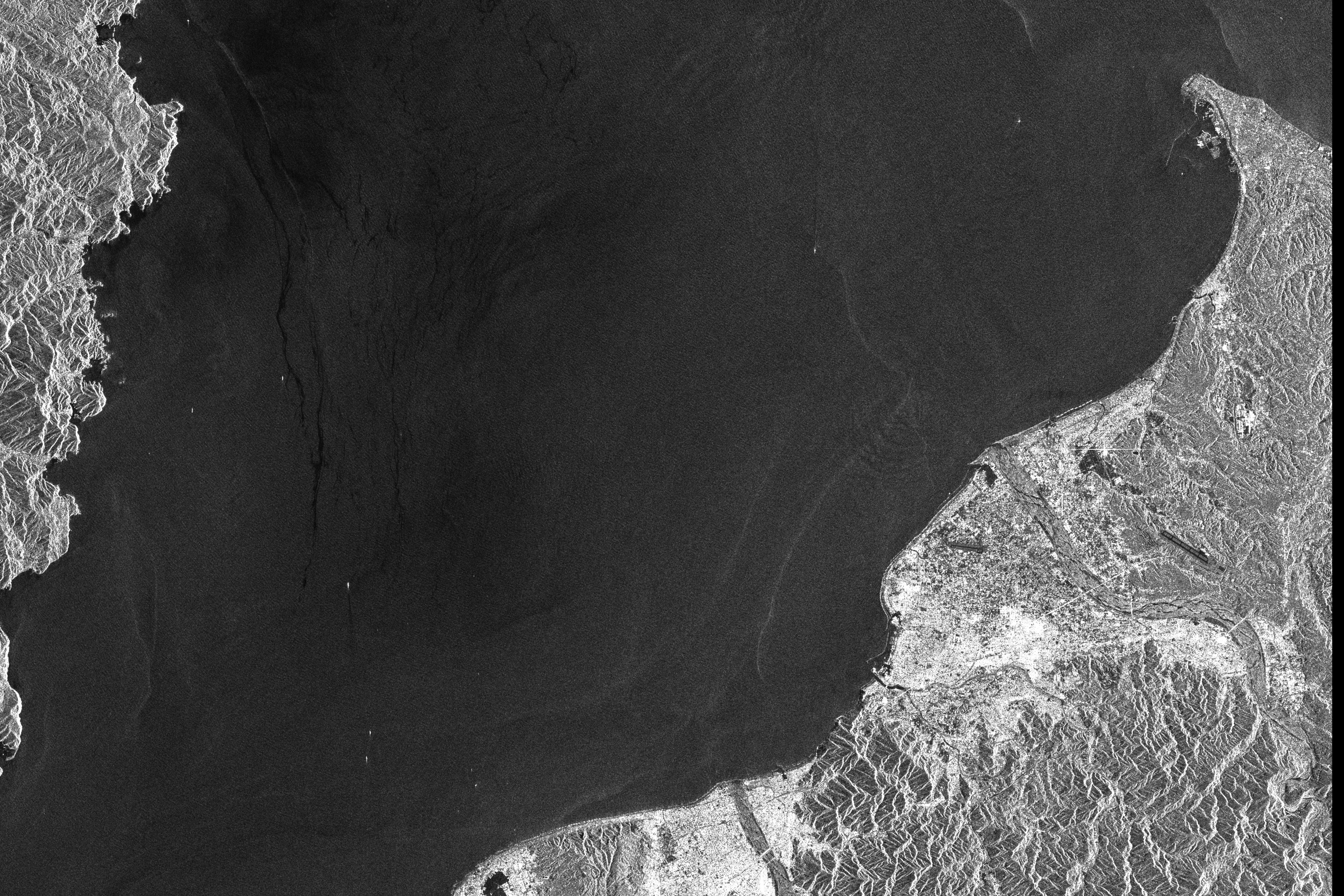}
    \caption{Example of a  SAR image \cite{LSSSDD}.}
    \label{fig:SARBEELD}
\end{figure}

\begin{figure}[h]
\centering
\begin{subfigure}{.5\textwidth}
  \centering
  \includegraphics[width=.5\linewidth,height=3cm]{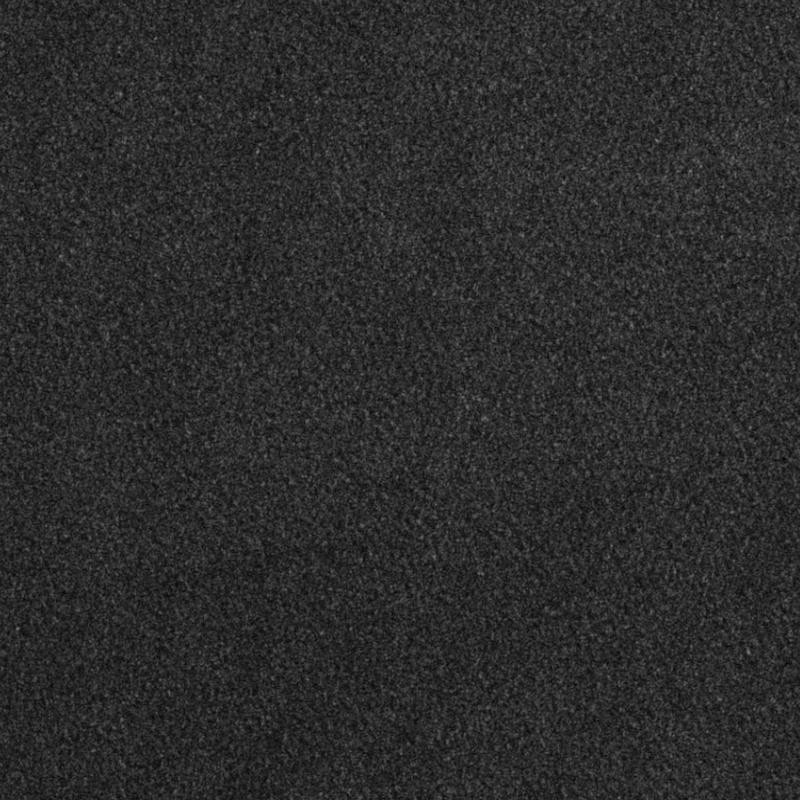}
  \caption{Sub-image without a ship.}
  \label{fig:geenschip}
\end{subfigure}%
\begin{subfigure}{.5\textwidth}
  \centering
  \includegraphics[width=.5\linewidth, height = 3cm]{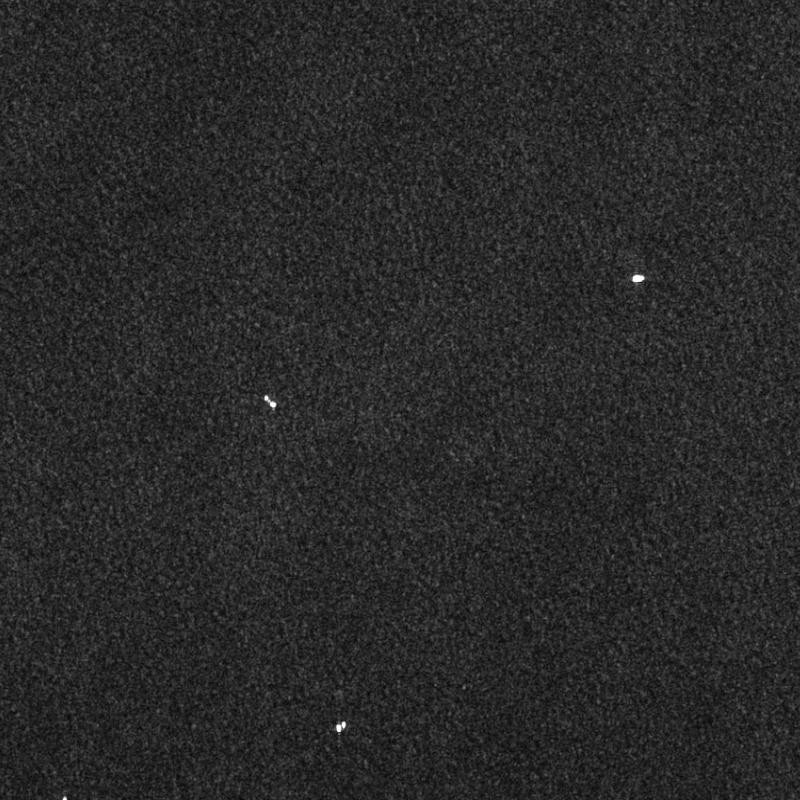}
  \caption{Sub-image with a ship.}
  \label{fig:schip}
\end{subfigure}%
\caption{ Example of a SAR sub-image without and with a ship.}
\label{fig:classificatievoorbeeld}
\end{figure}

In Figure \ref{fig:SARBEELD} an example of a SAR image is given.
Figure \ref{fig:classificatievoorbeeld} depicts a sub-image containing a ship and one without a ship.
Table \ref{tab:datasets} information about the split of the LSSSDD for training and testing is given.

\begin{table*}[h]
\addtolength{\tabcolsep}{-1pt}
\centering
\begin{tabular}{ | c ||c | c | c | c  | c  |}       
    \hline
    {Dataset} & {Ids} & {\# Subimgs} & {\# Vessels} & {\% Subimg with} & {\# Vessels / } \\ 
    {} & {} & {} & {} & {a vessel} & {subimg with a vessel} \\
    \hline
    \hline
    train & 1-10 & 6000 & 3637 & 18.7\% & 3.24 \\
    test & 11-15 & 3000 & 2378 & 24.5\% & 3.23 \\
    \hline
\end{tabular}%
\caption{Overview of the split into train and test set of the LSSSDD \cite{LSSSDD}.}
\label{tab:datasets}
\end{table*}

\newpage

\subsection{Patterns for the correlation algorithm}

\begin{figure}[htbp]
\centering
\begin{subfigure}{.45\textwidth}
  \centering
  \includegraphics[width=3cm]{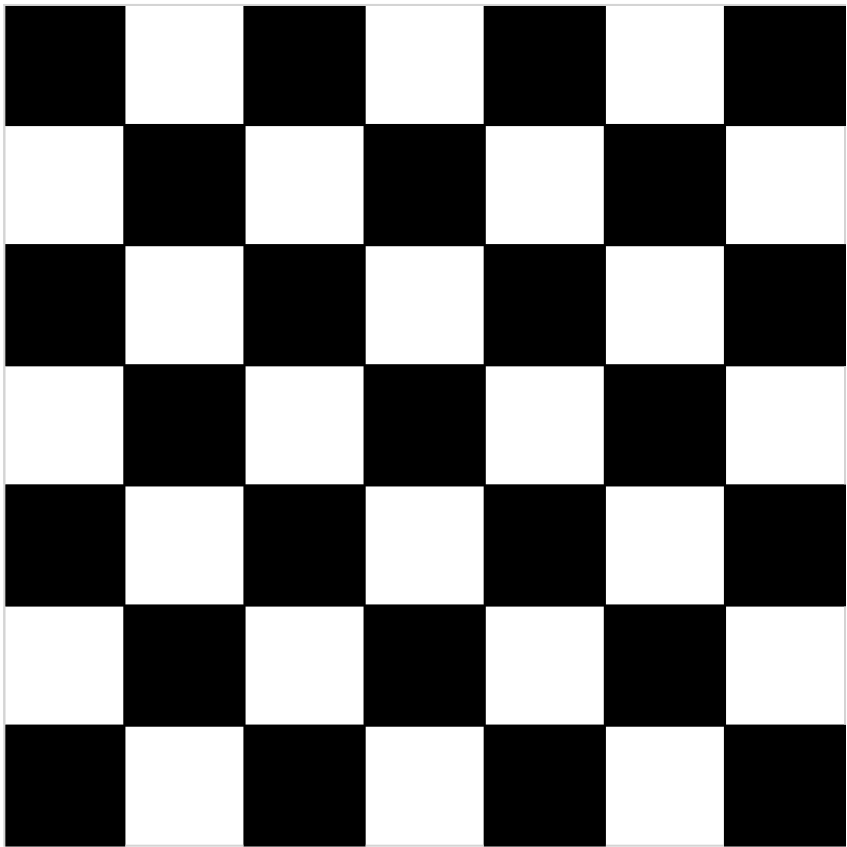}
  \caption{Checkers pattern.}
  \label{fig:dampatroon}
\end{subfigure}
\begin{subfigure}{.45\textwidth}
  \centering
  \includegraphics[width=3cm]{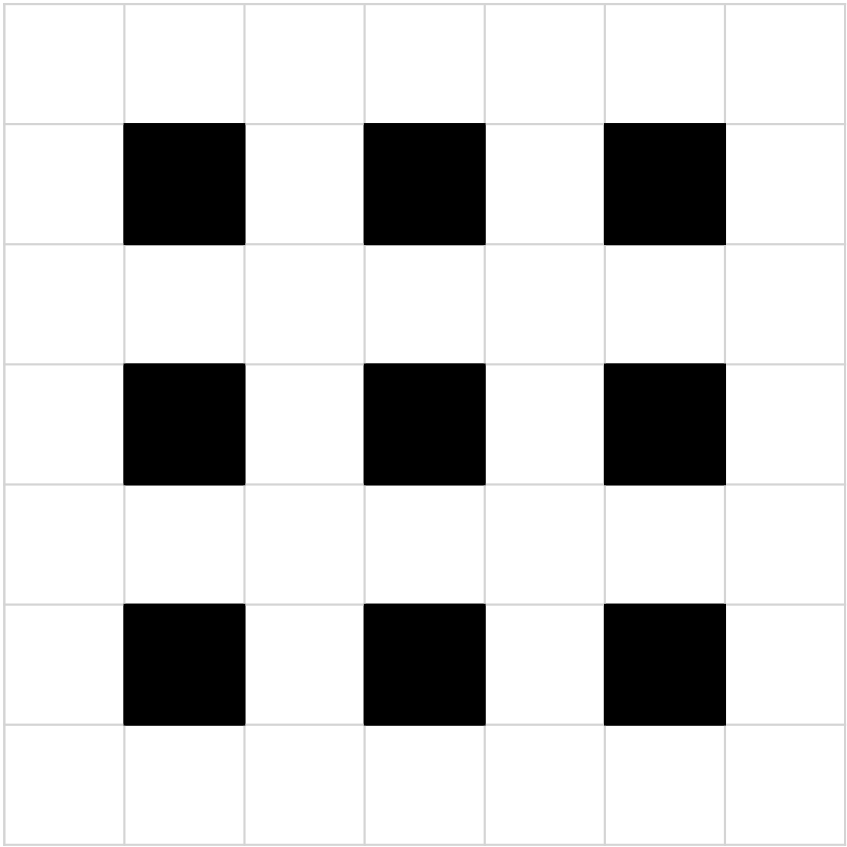}
  \caption{$\alpha$ pattern.}
  \label{fig:19patroon}
\end{subfigure}
\caption{Patterns used in the correlation-optimization.}
\label{fig:patronen}
\end{figure}

Figure \ref{fig:patronen} shows the two used patterns for the correlation algorithm.

\subsection{Results and $\digamma_{\beta}$ scores for the optimizations}

\begin{table*}[htbp]
\addtolength{\tabcolsep}{-1pt}
\centering
\begin{tabular}{| c  | c | c | c | c c | c | c| c |}       
\hline
    {Method} &{Precision} & {Recall} & {AP} & \multicolumn{2}{ c |}{RT}  & {$\digamma_{1}$}& {$\digamma_{0.5}$}& {$\digamma_{0.25}$} \\
    {} & {} & {} & {} & {GPU} & {CPU} & {} & {} & {} \\
    \hline
    \hline
    $f_\text{clf}(I,0)$	&	0.788	&	0.724	&	0.693	&	35\%	&	32\%	&	\textbf{0.673}	&	\textbf{0.684}	&	0.690	\\
    $f_\text{clf}(I,0.1)$	&	0.776	&	0.731	&	0.698	&	38\%	&	35\%	&	0.659	&	0.682	&	0.693	\\
    $f_\text{clf}(I,0.2)$	&	0.768	&	0.741	&	0.706	&	44\%	&	41\%	&	0.622	&	0.670	&	\textbf{0.695}	\\
    $f_\text{clf}(I,0.3)$	&	0.758	&	0.746	&	0.710	&	70\%	&	68\%	&	0.421	&	0.557	&	0.657	\\
    $f_\text{clf}(I,0.38)$	&	0.754	&	0.748	&	0.711	&	95\%	&	93\%	&	0.098	&	0.203	&	0.410	\\
    \hline
   
    \hline
\end{tabular}%
\caption{Results Faster R-CNN model with $f_\text{clf}$}
\label{tab:Detectieclassificatie}
\end{table*}

\begin{table*}[h]
\addtolength{\tabcolsep}{-1pt}
\centering
\begin{tabular}{| c |c | c |c | c c | c | c | c |}       
\hline
    {Method} &{Precision} & {Recall} & {AP} & \multicolumn{2}{ c |}{RT}  & {$\digamma_{1}$}& {$\digamma_{0.5}$}& {$\digamma_{0.25}$} \\
    {} & {} & {} & {} & {GPU} & {CPU} & {} & {} & {} \\
    \hline
    \hline
    $f_\text{cor-checkers}(I,3,[1,0.1,0.1],0.45)$	&	0.769	&	0.689	&	0.657	&	63\%	&	63\%	&	0.470	&	\textbf{0.566}	&	0.627	\\
    $f_\text{cor-checkers}(I,2,[1,0.33],0.5)$ 	&	0.768	&	0.672	&	0.640	&	62\%	&	62\%	&	\textbf{0.479}	&	0.564	&	0.616	\\
    $f_\text{cor-checkers}(I,2,[1,0.1],0.5)$ 	&	0.768	&	0.672	&	0.640	&	62\%	&	62\%	&	0.478	&	0.564	&	0.616	\\
    $f_\text{cor-checkers}(I,1,[1],0.5)$ &	0.766	&	0.508	&	0.701	&	67\%	&	67\%	&	0.444	&	0.556	&	0.630	\\
    $f_\text{cor-checkers}(I,2,[1,0.1],0.375)$ 	&	0.766	&	0.701	&	0.667	&	67\%	&	67\%	&	0.442	&	0.554	&	0.629	\\
    $f_\text{cor-checkers}(I,3,[1,0.1,0.1],0.35)$ 	&	0.764	&	0.706	&	0.672	&	68\%	&	68\%	&	0.436	&	0.552	&	\textbf{0.632}	\\
    $f_\text{cor-checkers}(I,3,[1,0.1,0.1],0.2)$  & 	0.760	&	0.735	&	0.699	&	80\%	&	80\%	&	0.315	&	0.470	&	0.611	\\
    $f_\text{cor-checkers}(I,3,[1,0.33,0.1],0.2)$ 	&	0.761	&	0.737	&	0.701	&	79\%	&	79\%	&	0.319	&	0.474	&	0.614	\\
    $f_\text{cor-checkers}(I,3,[1,0.5,0.25],0.2)$ 	&	0.761	&	0.738	&	0.703	&	81\%	&	81\%	&	0.300	&	0.457	&	0.607	\\
    
    \hline
  
\end{tabular}%
\caption{Results Faster R-CNN model with $f_\text{cor-checkers}$.}
\label{tab:fincorrelatiedam}
\end{table*}

\begin{table*}[h]
\addtolength{\tabcolsep}{-1pt}
\centering
\begin{tabular}{| c | c | c | c |c c| c| c | c |}       
\hline
    {Method} &{Precision} & {Recall} & {AP} & \multicolumn{2}{ c |}{RT}  & {$\digamma_{1}$}& {$\digamma_{0.5}$}& {$\digamma_{0.25}$} \\
    {} & {} & {} & {} & {GPU} & {CPU} & {} & {} & {} \\
    \hline
    \hline
    $f_\text{cor-$\alpha$}(I,3,[1,0.1,0.1],0.35)$ &	0.778	&	0.660	&	0.631	&	53\%	&	53\%	&	0.536	&	0.589	&	0.618	\\
    $f_\text{cor-$\alpha$}(I,2,[1,0.1],0.4375)$ 	&	0.788	&	0.644	&	0.616	&	50\%	&	50\%	&	\textbf{0.553}	&	0.589	&	0.608	\\
    $f_\text{cor-$\alpha$}(I,3,[1,0.5,0.1],0.35)$ 	&	0.775	&	0.651	&	0.623	&	52\%	&	52\%	&	0.545	&	\textbf{0.589}	&	0.612	\\
    $f_\text{cor-$\alpha$}(I,3,[1,0.1,0.1],0.2)$ &	0.764	&	0.694	&	0.662	&	62\%	&	62\%	&	0.486	&	0.578	&	0.635	\\
    $f_\text{cor-$\alpha$}(I,2,[1,0.1],0.1875)$ 	&	0.770	&	0.690	&	0.658	&	60\%	&	60\%	&	0.496	&	0.582	&	0.634	\\
    $f_\text{cor-$\alpha$}(I,2,[1,0.1],0.125)$ 	&	0.767	&	0.696	&	0.664	&	62\%	&	62\%	&	0.487	&	0.580	&	\textbf{0.637}	\\
    $f_\text{cor-$\alpha$}(I,3,[1,0.5,0.1],0.1)$ &	0.757	&	0.743	&	0.708	&	79\%	&	79\%	&	0.327	&	0.479	&	0.614	\\
    $f_\text{cor-$\alpha$}(I,3,[1,0.75,0.1],0.1)$ 	&	0.757	&	0.743	&	0.708	&	81\%	&	81\%     & 0.304	&	0.460	&	0.606	\\
    $f_\text{cor-$\alpha$}(I,3,[1,1,0.1],0.1)$ &	0.757	&	0.743	&	0.708	&	81\%	&	81\%	&	0.304	&	0.459	&	0.606	\\
    \hline

\end{tabular}%
\caption{Results Faster R-CNN model with $f_\text{cor-$\alpha$}$.}
\label{tab:fincorrelatie1_9}
\end{table*}

\begin{table*}[h]
\addtolength{\tabcolsep}{-1pt}
\resizebox{\columnwidth}{!}{%
\centering
\begin{tabular}{|c | c | c  c | c | c | c |}     
\hline
    {Method} & {AP} & \multicolumn{2}{ c |}{RT}  & {$\digamma_{1}$}& {$\digamma_{0.5}$}& {$\digamma_{0.25}$} \\
    {}  & {} & {GPU} & {CPU} & {} & {} & {} \\
    \hline
    \hline
    $f_\text{cor-$\alpha$}(I,2,[1,0.5],0.25)$ \text{ en } $f_\text{clf}(I,0)$	&	 0.638 	&	25\%	&	24\%	&	\textbf{0.688}	& 0.657 & 	0.643 \\
    $f_\text{cor-$\alpha$}(I,2,[1,0.5],0.25)$ \text{ en }	$f_\text{clf}(I,0.2)$ &	 0.664 	&	32\%	&	30\%   & 0.671 & 	\textbf{0.666} & 	\textbf{0.665} \\
    \hline
\end{tabular}%
}
\caption{Results Faster R-CNN model with $f_\text{cor}$ and $f_\text{clf}$}
\label{tab:Classcorr}
\end{table*}

\newpage
\subsection{Hyperparameters for the used DL models}

\begin{table}[h]
\addtolength{\tabcolsep}{-1pt}
\centering
\begin{tabular}{ c c}       
    \toprule
    {Hyperparameter} & {Value} \\ 
    \midrule
    Batch size & 256 \\
    Base learning rate & 0.001 \\
    Momentum beta & 0.9 \\
    Weight decay & 0.0001 \\
    Anchor sizes & 10, 16, 32, 40, 64 \\
    Anchor aspect ratio's & 0.5, 1, 2 \\
    NMS threshold & 0.5 \\
    \bottomrule
\end{tabular}
\caption{Hyperparameters for the Faster R-CNN model}
\label{tab:hpfasterrcnn}
\end{table}
\begin{table}[h!]
\addtolength{\tabcolsep}{-1pt}
\centering
\begin{tabular}{ c c}       
    \toprule
    {Loss function} & {Function} \\ 
    \midrule
    RPN classification loss function & softmax binary CEL \\
    RPN localisation loss function & L1 loss \\
    Bounding box localisation loss function & smooth L1 loss \\
    Bounding box classification loss function & softmax CEL \\
    
    \bottomrule
\end{tabular}
\caption{Loss functions for the Faster R-CNN model}
\label{tab:lffasterrcnn}
\end{table}
\begin{table}[h!]
\addtolength{\tabcolsep}{-1pt}
\centering
\begin{tabular}{ c c}       
    \toprule
    {Hyperparameter} & {Value} \\ 
    \midrule
    Batch size & 32 \\
    Base learning rate & 0.0001 \\
    \bottomrule
\end{tabular}
\caption{Hyperparameters for the MobileNetV2 model}
\label{tab:hpfasterrcnnn}
\end{table}
\end{document}